\documentclass[lettersize,journal]{IEEEtran}
\usepackage{amsmath,amsfonts}
\usepackage{algorithmic}
\usepackage{algorithm}
\usepackage{array}

\usepackage{textcomp}
\usepackage{stfloats}
\usepackage{url}
\usepackage{verbatim}
\usepackage{graphicx}
\usepackage{cite}
\usepackage{multicol}
\usepackage{enumitem}
\usepackage{multirow}

\usepackage{color}
\usepackage{soul}

\usepackage[caption=false,font=footnotesize]{subfig}

\usepackage[capitalise]{cleveref}

\begin{document}

\title{Worker Activity Recognition in Manufacturing Line Using Near-body Electric Field}

\author{Sungho~Suh,~\IEEEmembership{Member,~IEEE,}
		Vitor Fortes Rey,
        Sizhen Bian,~\IEEEmembership{Member,~IEEE,}
        Yu-Chi Huang,
        Jože M. Rožanec,~\IEEEmembership{Member,~IEEE,}
        Hooman Tavakoli Ghinani,
        Bo Zhou,
        Paul~Lukowicz,~\IEEEmembership{Member,~IEEE}
		\IEEEcompsocitemizethanks{\IEEEcompsocthanksitem S. Suh, V. F. Rey, S. Bian, Y. Huang, H. T. Ghinani, B. Zhou, and P. Lukowicz are with German Research Center for Artificial Intelligence (DFKI), 67663 Kaiserslautern, Germany. S. Suh, V. F. Rey, Y. Huang, B. Zhou, and P. Lukowicz are also with the Department of Computer Science, RPTU Kaiserslautern-Landau, 67663 Kaiserslautern, Germany. J. M. Rožanec is with Jožef Stefan International Postgraduate School, Ljubljana, Slovenia.\protect
			% note need leading \protect in front of \\ to get a newline within \thanks as
			% \\ is fragile and will error, could use \hfil\break instead.
        \IEEEcompsocthanksitem Corresponding Author: Sungho Suh (Sungho.Suh@dfki.de)
		}
        % <-this % stops a space
% \thanks{This paper was produced by the IEEE Publication Technology Group. They are in Piscataway, NJ.}% <-this % stops a space
% \thanks{Manuscript received April 19, 2021; revised August 16, 2021.}
}

% The paper headers
\markboth{IEEE Internet of Things Journal}%
{Suh \MakeLowercase{\textit{et al.}}: Worker Activity Recognition in Manufacturing Line using Near-body Electric Field}

% \IEEEpubid{0000--0000/00\$00.00~\copyright~2021 IEEE}
% Remember, if you use this you must call \IEEEpubidadjcol in the second
% column for its text to clear the IEEEpubid mark.

\maketitle

\begin{abstract}
Manufacturing industries strive to improve production efficiency and product quality by deploying advanced sensing and control systems. Wearable sensors are emerging as a promising solution for achieving this goal, as they can provide continuous and unobtrusive monitoring of workers' activities in the manufacturing line. This paper presents a novel wearable sensing prototype that combines IMU and body capacitance sensing modules to recognize worker activities in the manufacturing line. To handle these multimodal sensor data, we propose and compare early, and late sensor data fusion approaches for multi-channel time-series convolutional neural networks and deep convolutional LSTM. We evaluate the proposed hardware and neural network model by collecting and annotating sensor data using the proposed sensing prototype and Apple Watches in the testbed of the manufacturing line. Experimental results demonstrate that our proposed methods achieve superior performance compared to the baseline methods, indicating the potential of the proposed approach for real-world applications in manufacturing industries. Furthermore, the proposed sensing prototype with a body capacitive sensor and feature fusion method improves by 6.35\%, yielding a 9.38\% higher macro F1 score than the proposed sensing prototype without a body capacitive sensor and Apple Watch data, respectively.
\end{abstract}


\section{Introduction}
    \label{sec:intro}
    
    % General Human activity recognition
    \IEEEPARstart{H}{uman} activity recognition (HAR) is a classification task for recognizing human movement. With the rapid development of wearable and mobile devices, HAR in ubiquitous computing environments has become a more important topic in better understanding people's daily activities and interactions with their living environments, which is studied extensively for the Internet of Things (IoT). HAR approaches have been researched for human movement classification and widely used in health, and medical care \cite{bachlin2009wearable, xu2018geometrical}, sports activities \cite{direkoglu2012team, margarito2015user, bian2019passive}, smart home \cite{roy2011possibilistic, wen2016adaptive}, and many other applications. 
    
    % A brief explanation of worker activity recognition in manufacturing line for worker’s safety
    Also, HAR using wearable sensors can be applied in manufacturing environments along with the recent growing interest in smart manufacturing and Industry 4.0 \cite{shrouf2014smart, wang2016implementing, rovzanec2022human}. In the manufacturing line, HAR can be utilized to quantify and evaluate worker performance \cite{tao2018worker}, understand workers' operational behavior \cite{aehnelt2014using}, support workers' operations with industrial robots \cite{roitberg2014human, roitberg2015multimodal}, and execute maintenance work \cite{ward2006activity}. In particular, activity recognition for worker safety in the manufacturing line is becoming increasingly important. Activity recognition, if performed reliably, would provide the ability to quickly identify workers' needs for assistance or prevent industrial accidents. 
    
    % Description of the SmartFactoryKL testbed use case
    In the present research, we deploy and test HAR using wearable sensors for a dedicated use case in a smart factory testbed where workers and mobile robots operate in the same environment to predict worker's movement intention to plan an optimal mobile robot route without collision risks between the worker and the robot. In the industrial environment, some workflow processes are done by workers for production reconfiguration and maintenance. It is important to predict human behavior to configure and optimize the mobile robot later to avoid possible collisions and to create safety zones. In the testbed, we assume that there is a mobile robot and a worker to check the industrial machine. The use case aims to maintain a high production level and worker safety during mobile robot activity.
        
    % The existing wearable sensor frameworks and limitations 
    In recent years, with the popularity of deep learning and micro-electromechanical technologies, wearable sensors have been introduced into the applications of HAR using multichannel time-series, for instance, measurements from inertial measurement unit (IMU) which combines accelerometer, gyroscope, and magnetometer \cite{ronao2015deep, yang2015deep, ha2016convolutional, ordonez2016deep, suh2022adversarial, suh2023tasked}. Distinguished from traditional machine learning methods, deep learning-based approaches make extracting and classifying the principle features from complex data with numerous sensor sources more convenient. Ronao et al. \cite{ronao2015deep} and Yang et al. \cite{yang2015deep} utilized Convolutional Neural Networks (CNN) to perform feature extraction with sliding windows on the multichannel IMU data. Ordonez and Roggen \cite{ordonez2016deep} proposed a Deep Convolutional Long Short-Term Memory (DeepConvLSTM), which used CNN and Long Short-Term Memory networks (LSTM) to learn features from the time-series multichannel IMU signals. Even though IMU has been dominantly utilized for HAR and has provided high performance, IMU still has limitations in tracking human activities as it can only measure movements as captured by the limb to which the sensor is attached. However, finding a competitive sensing modality other than IMU is hard.

    % The existing wearable sensor work using capacitive sensors (After writing Related Work Section)
    To provide another option of the sensing modality for HAR and address IMU's limitations, the body capacitance, which captures the electric field feature between the body and the environment has been proposed and utilized \cite{aliau2012fast, bian2022systematic}. Researchers have looked into this novel property to extend wearable motion-sensing ability, as it can capture both body movement and environment variations, such as the intrusion of a second body and the approaching of surrounding grounded objects. This approach enjoys the benefits of low cost and low power consumption, which is critical for wearable devices. Existing works, such as \cite{staudt_pascal_2022_6798242} and \cite{canat2016sensation}, demonstrate the potential of body capacitance sensing for distance measurement and touch gesture recognition. By using a customized wearable capacitive sensing front-end, we aim to extend the ability of IMU-based activity recognition.
    
    % A brief statement of the proposed hardware design (After writing Hardware Sensor Design Section)
    % limitations of feature fusion in the existing works 
    % A brief statement of the proposed feature fusion model for IMU and capacitive sensor data
    This paper presents a novel wearable sensing prototype that combines IMU and body capacitance sensing modules to recognize worker activities in the manufacturing line. To handle these multimodal sensor data, we propose and compare early and late sensor data fusion approaches for the multi-channel time-series convolutional neural networks (MC-CNN) and the DeepConvLSTM. The early data fusion approach merges the IMU and capacitive sensor data at the raw data level and processes the merged data using the neural network models, while the late feature fusion approach processes each sensor data separately using a feature extractor and concatenates the extracted features to recognize the worker activities in manufacturing line using the classifier. First, to evaluate the proposed hardware and neural network model, we collected and annotated sensor data using the proposed sensing prototype and Apple Watches in the testbed of the manufacturing line. Then, through a series of experiments on the collected datasets, we demonstrate the effectiveness of the proposed wearable sensors and neural network model. 
    
    %The contributions of our work
    The contributions of this paper are: 
    \begin{itemize} 
    \item we propose a wearable sensor design with the IMU and capacitive sensors for human activity recognition in the manufacturing line.
    \item we adopt a feature fusion method to combine the multimodal wearable sensors to improve HAR performance. 
    \item to validate the proposed hardware sensor design and the feature fusion model, we collected and annotated sensor data in the testbed of the manufacturing line. 
    \end{itemize}
    % 1. We propose a wearable sensor design with IMU and the capacitive sensor for worker activity recognition in the manufacturing line for worker safety.
    % 2. Comparison experiments between the proposed sensor hardware and Apple Watch with deep learning-based activity recognition modules on the sensor data collected at the manufacturing testbed
    % 3. The use case for worker safety in manufacturing line itself.
    
    % Paper structure
    This paper is structured as follows: \cref{sec:relatedwork} presents related work, and \cref{sec:hardware} provides the proposed hardware sensor design details. \cref{sec:ExpDesign} introduces the use case at the testbed, experimental design, and the neural network models with the feature fusion approach. \cref{sec:ExpResults} shows the experimental results on the collected sensor data by the designed sensor and Apple Watch. \cref{sec:conclusion} concludes this paper and outlines future work.
    
    \section{Related Work}
    \label{sec:relatedwork}

    % IoT assisted factory
    % Worker activity recognition in manufacturing line
    \subsection{Internet of Things (IoT)-assisted Smart Factory}
    With the recent advances in sensing technologies, recognizing human activities using various sensors such as IMU sensors \cite{yang2015deep, ha2016convolutional, ordonez2016deep, suh2022adversarial}, microphones \cite{korpela2015evaluating, bello2020facial}, cameras \cite{pirsiavash2012detecting}, and magnetic sensors \cite{maekawa2013activity, pirkl2015mbeacon} has become more feasible. While supervised machine learning techniques have been widely used for recognizing activities using labeled training data \cite{lukowicz2004recognizing}, researchers have also explored methods that reduce the effort required for data collection. Among the popular approaches, transfer learning, activity modeling, and clustering techniques are employed.
    
    In line with the growing interest in smart manufacturing and Industry 4.0 \cite{radziwon2014smart}, studies on recognizing and supporting factory work using sensor technologies \cite{tao2018worker, al2019action, qingxin2019unsupervised, al2022fusing} have gained significant attention. Several studies have explored the use of sensors for activity recognition in the manufacturing area. Koskimaki et al. \cite{koskimaki2009activity} used a wrist-worn IMU sensor to capture arm movements and a K-Nearest Neighbors model to classify five activities for industrial assembly lines. Using machine learning models, Ward et al. \cite{ward2005gesture} utilized acceleration and audio data from wrist-worn devices to recognize woodworking activities. Stiefmeier et al. \cite{stiefmeier2006combining} utilized ultrasonic and IMU sensors to identify worker activities in a bicycle maintenance scenario using a Hidden Markov Model classifier. They later proposed a string-matching-based segmentation and classification method using multiple IMU sensors to recognize worker activities in car manufacturing tasks \cite{stiefmeier2007fusion, stiefmeier2008wearable}. Maekawa et al. \cite{maekawa2016toward} proposed an unsupervised method for lead time estimation of factory work using signals from a smartwatch with an IMU sensor. These studies demonstrate the potential of sensors in recognizing worker activities in manufacturing environments, especially using machine learning techniques.

    \subsection{Sensor-based Human Activity Recognition}
    % Activity recognition with capacitive sensors
    The IMU currently dominates wearable sensor-based long-term human activity recognition. It has been embedded in almost every portable smart device, benefiting from its low cost, low power consumption, and tiny form factor. However, we noticed that the ability of IMU is limited to the sensing object, namely only the moving pattern of the wearer. In a manufacturing environment, the body-environment and body-machine interaction play critical roles in security. 
    % Existing works using Apple Watch?

    To extend the wearable motion-sensing ability, researchers explored a novel natural property, the body capacitance \cite{bian2019wrist, bian2022systematic, aliau2012fast}, which describes the electric field between the body and the environment. The benefit of using body capacitance as a competitive sensing source is that the body movement and environment variation, like the contact or proximity of a second conductive entity to the surrounding grounded object, will cause efficient signal variation for pattern recognition. An example work is \cite{staudt_pascal_2022_6798242}, where the authors installed the transceivers on multiple subjects, making the body part of an electric system that works regardless of the location of the measurement and orientation of the body in the space. Because of the inter-body capacitance, the transmitted electric field from the transmitter can be picked up by the receiver when the two bodies are close. The received signal is then used to deduce the distance information and generate the sound. In \cite{canat2016sensation}, the authors designed a piece of equipment for detecting touch patterns (touch with one-Finger, bro-fist, palm Touch, four-Finger) between players and introduced the game, which is a collaborative game designed to be played with the social touch. The background of such a device is based on the body capacitance relationship among users and different touch gestures will generate different capacitance variation patterns, which are then used for touch gesture recognition as the game's input. In summary, the body capacitive signal provides an alternative approach for wearable motion and environmental variation sensing and enjoys the benefits of low cost and low power consumption, which is critical for wearable devices. By using a customized wearable capacitive sensing front-end, we extend the ability of IMU-based activity recognition.

    % ML models in wearable activity recognition
    \subsection{ML models in wearable activity recognition}

Machine learning methods provide a mechanism to accurately learn and detect human activities based on sensor data. Numerous studies have shown that human activity recognition strongly depends on the amount, modality, and placement of sensors considered \cite{janidarmian2017comprehensive}. While the placement is critical to accurately measuring certain activities, collecting information from multiple and different sensors has shown to be an effective method to reduce uncertainty and enhance the reliability of the overall HAR system \cite{nweke2019data}. To use sensory data, the streams are usually divided into subsequent streams considering a fixed-size (overlapping or non-overlapping) sliding window. Furthermore, multiple approaches have been developed to perform information fusion. Among frequently used machine learning algorithms for human activity recognition, we find support vector machines, k-nearest neighbors, and artificial neural networks \cite{qiu2022multi}. In particular, Hassan et al. \cite{hassan2018robust} compared using Deep Belief Networks, Support Vector Machines, and Artificial Neural Networks for activity recognition. Zhou et al. \cite{zhou2015motion} compared Artificial Neural Networks with Stacked Autoencoders, considering the latest had a better discriminative capability. Conversely, Islam et al. \cite{islam2019evaluation} explored transfer learning to mitigate the lack of labeled data and compared the approach to a Convolutional Neural Network. In the same line, Feng et al. \cite{feng2019few} leverage transfer learning in a few-shot learning setting achieving state-of-the-art results. Among other methods and complementary approaches, we can mention the use of many deep learning architectures (e.g., recurrent neural networks, autoencoders (sparse, denoising, or variational), generative adversarial networks), and approaches such as active learning \cite{wang2019deep,chen2021deep,gu2021survey}.

\section{Hardware Sensor Design} % Sizhen
    \label{sec:hardware}

    \begin{figure}[!t]
        \centering
        \includegraphics[width=\columnwidth]{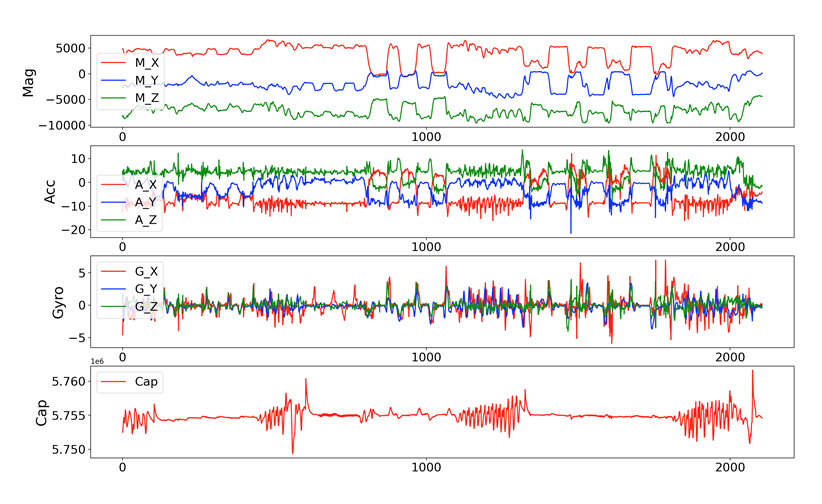}
        \caption{Example of recorded sensor signals from the sensing prototype}
        \label{fig:examplesignal}
    \end{figure} 
    \begin{figure}[!t]
        \centering
        \subfloat[]{\includegraphics[width=0.45\columnwidth, height = 3.4cm]{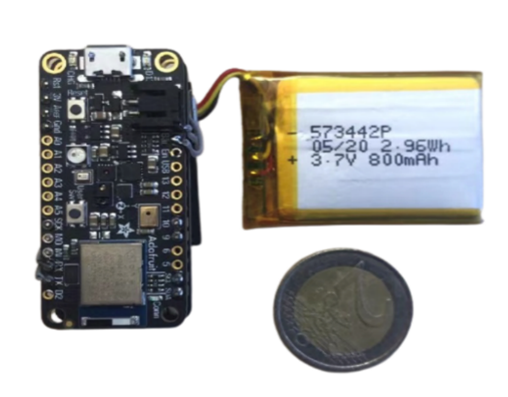}}
        \subfloat[]{\includegraphics[width=0.30\columnwidth, height = 3.4cm]{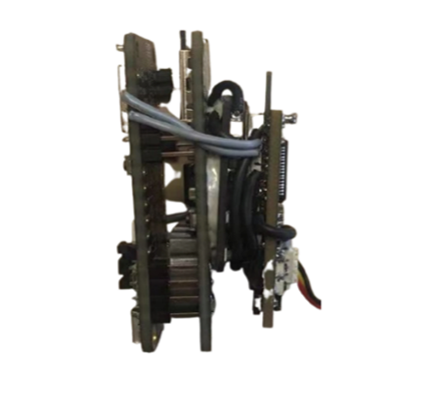}}
        %\hfill
        \subfloat[]{\includegraphics[width=0.25\columnwidth, height = 3.4cm]{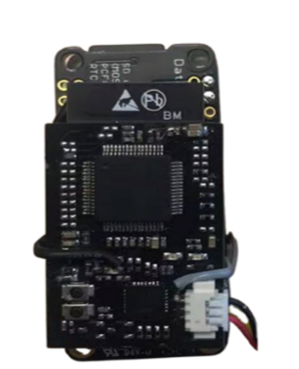}}
        %\subfloat[]{\includegraphics[width=0.35\columnwidth]{Figures/prototype_d.png}}
        \caption{The combined sensing prototype for worker's activity recognition: (a) top view: the motherboard and the battery; (b) side view: the stacked three separate boards: motherboard, capacitive sensing board, and the data logging board; (c) bottom view: the front-end of the capacitive sensor, mainly a high-resolution, low noise analog-to-digital converter, as described in \cite{bian2019passive}.%; %(d) the prototype worn on the wrist, with an electrode attached to the skin, since the capacitive sensor senses the body potential to indicate the body capacitance variation.
        }
        \label{F:prototype}
    \end{figure} 
    
    We designed a wearable sensing prototype with the traditional IMU sensing module and body capacitance sensing module to track human activity in the manufacturing line. The human body has good electric properties like skin conductivity, and the body capacitance sensing module could have significant potential to interact with the environments, especially the ground. Even though the person wears clothes, the relationship between the human body and the environment can be interpreted because the body is a pure conductive plate and can act as a natural capacitor. We explored the physiological signals from the IMU and body capacitance sensors. We evaluated our human activity recognition module for predicting user movement trajectories by designing a sensing prototype that collected real-life data. \cref{fig:examplesignal} depicts the signals of the four low-power motion sensors used in the prototype as the subject repetitively performed actions in an office environment, such as opening doors and windows, touching the desktop, and interacting with a floor lamp. The four sensors included the traditional inertial measurement unit (IMU) consisting of an accelerometer, gyroscope, and magnetometer, and a novel capacitive sensor that indirectly reflects changes in body capacitance during movement or interactions with the environment, which is competitive in recognizing certain body activities. Previous studies have demonstrated how this novel body capacitance signal contributes to recognizing individual and collaborative human activities \cite{bian2022using}. Our sensing prototype combined three boards: an nRF52840 backend motherboard from Adafruit Feather, a customized human body capacitance sensing board, and a data logger board, which recorded sensor data at a rate of 25 Hz. Both the IMU and body capacitance sensors consumed sub-milliwatts of power and cost less than a few dollars, making them ideal for pervasive and wearable motion sensing.

    We used the sensing prototype and simple neural networks to evaluate our human activity recognition module and predict user movement trajectories based on artificial intelligence models. The customized body capacitance board was verified to sense body movement and environmental context by measuring skin potential signals. To avoid data loss in the factory environment full of 2.4G Hz wireless signals, such as WiFi and Bluetooth, we recorded the sensor data locally in an SD card and finally synchronized it by checking some predefined actions. The sensing component, including the IMU and body capacitance sensor, consumed sub-mW power, and we used a 3.7V chargeable lithium battery for the power supply. \cref{F:prototype} shows the different views of the combined sensing prototype. While the module's connectivity was not used during data collection, the hardware has wireless WiFi and Bluetooth connectivity which can be used in the end application.

    \section{Worker Activity Recognition in Manufacturing Line}
    \label{sec:ExpDesign}

     \subsection{Use Case at the SmartFactory Testbed} % Hooman
     
            \begin{figure*}[!t]
                \centering
                \includegraphics[width=0.9\textwidth]{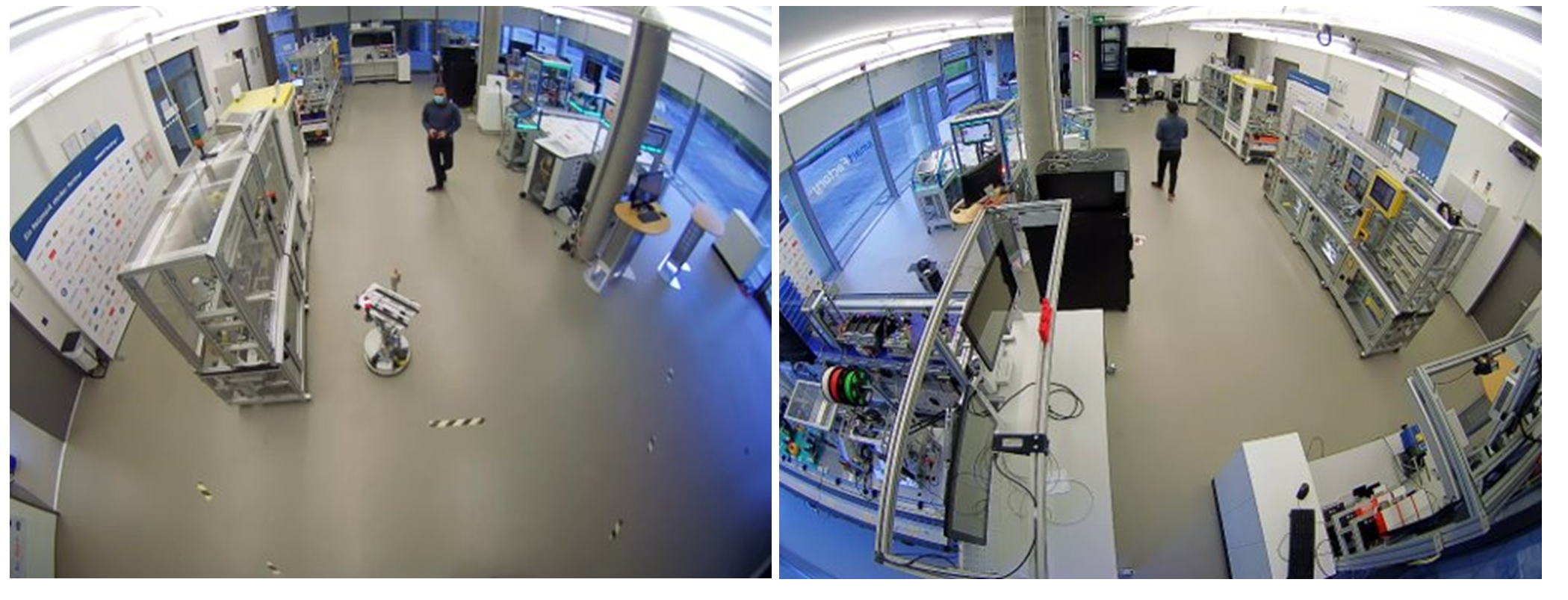}
                \caption{The top view of the SmartFactory from two ceiling cameras installed on the two opposite corners. The modules, workers, and robot are illustrated. In the testbed, the human and robots share the working areas which increase the demand for safety and reliability.}
                \label{fig:f_pilot}
            \end{figure*}
            \begin{figure*}[!t]
                \centering
                \includegraphics[width=0.9\linewidth]{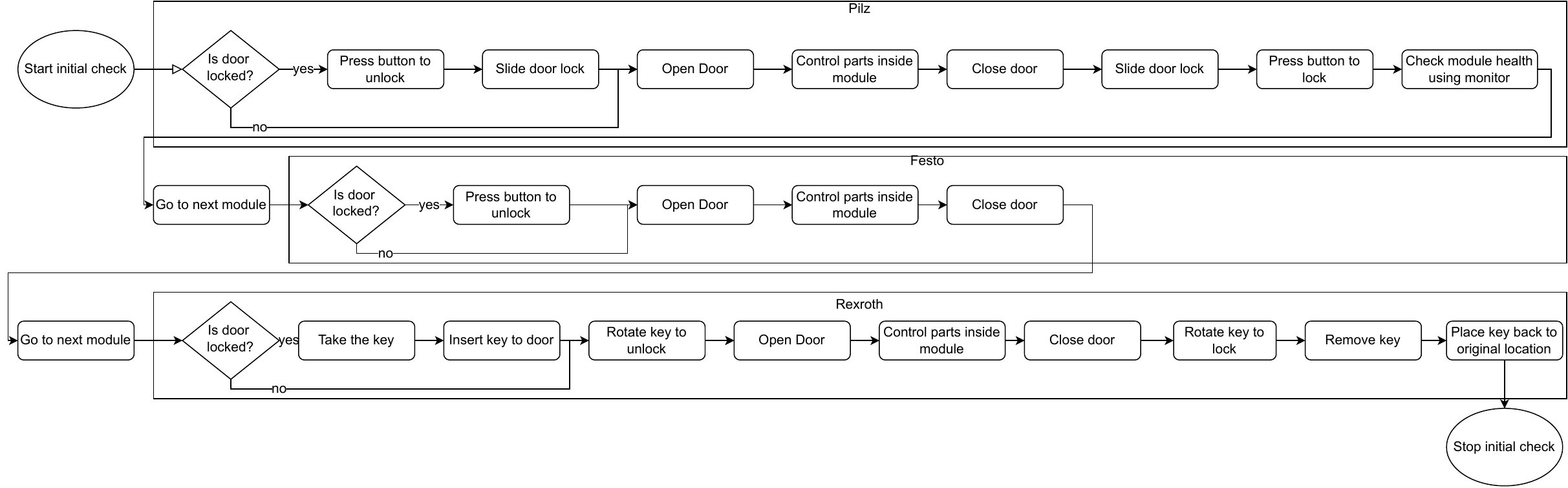}
                \caption{A sample of the workflow of the worker activities conducted during the inspection in SmartFactory.}
                \label{fig:workflow}
            \end{figure*} 
            
         Technology Initiative SmartFactory KL (SmartfactoryKL) as a non-profit organization is focusing on developing the next generation of manufacturing with the industry specialist called the factory innovators. This organization consists of the department of machine tools and controls (WSKL) at the TU Kaiserslautern and the Innovative Factory Systems research unit (IFS) at the German Research Center for Artificial Intelligence (DFKI). 
         
         Thanks to the two worldwide unique manufacturer-independent demonstrations of the SmartFactory, it is conceivable to develop, test and deploy innovative ICT technologies in a realistic industrial production environment. The demonstrator ecosystem will play an increasing role as a test environment in the coming years for the European GAIA-X sub-project SmartMA-X. These flexible production systems can be positioned in highly individual arrangements and make the industrial environment more dynamic.
    
         In this research, the SmartFactory is considered as the testbed to deploy and test the Human action recognition module in a dedicated use case called "Human action recognition and prediction in the respective environment."(See \cref{fig:f_pilot}). To come up with this scenario, we defined the worker activity pipeline (worker presence in the pilot and collaboration with different modules and robots with 20 diverse activities (see \cref{fig:workflow})). The prediction of human action in the production lines is essential when the worker is facing the moving robot. Whether in collaboration with a robot or in the path generation of the Robot to preclude any possible collision in layout. In other words, relying on the human location and his/her next possible action, the moving robot's path in the environment can be manipulated to minimize any possible collision which can be happened and would demolish the safety and reliability of the industrial environment, especially for human presence.

    \subsection{Data Recording} 
        To evaluate the performance of the proposed sensor hardware incorporating an IMU and capacitive sensors, we conducted an experiment in which twelve volunteers wore the proposed wearable sensing prototype and an Apple Watch while performing various tasks designed to simulate typical worker scenarios during normal daily work. The tasks included opening and closing doors, walking, checking parts inside a module, and interacting with a touch screen. The experiment was divided into five sessions, each lasting between 2-3 minutes. Some sessions were conducted and recorded in a different direction from the flow chart to ensure that the results were robust. All participants signed an agreement following the policies of the university's committee for the protection of human subjects, and the experiment was video recorded for further confidential analysis including ground truth activity annotation. The observer and the participant followed an ethical/hygienic protocol according to public health guidelines.
        
        \begin{figure}[!t]
            \centering
            \subfloat[]{\includegraphics[width=\columnwidth]{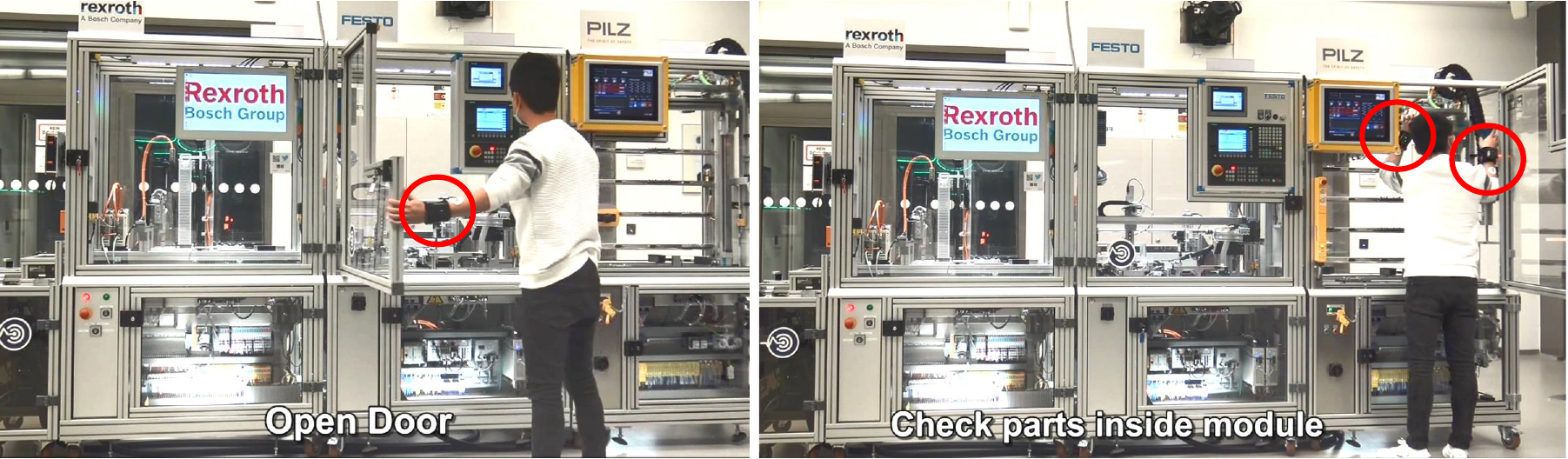}}
            \hfill
            \subfloat[]{\includegraphics[width=\columnwidth]{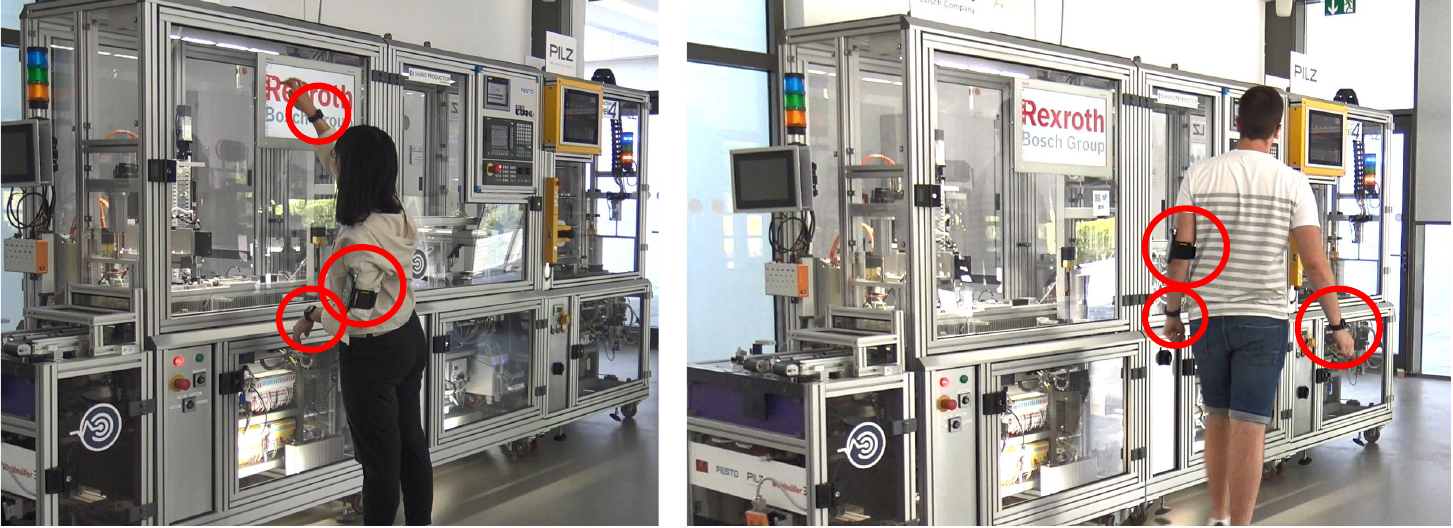}}
            \caption{The wearable sensors attached on the participants. (a) The proposed wearable sensing prototype on both wrists, (b) Apple Watches on both wrists and iPhone mini on left arm. }
            \label{fig:datarecording}
        \end{figure}
        
        The proposed prototype wearable sensors were attached to both wrists of the participants, while the Apple Watches were attached to both wrists and an iPhone mini was attached to the left arm. The example images of the data recording with the proposed prototype sensors and the Apple Watches are depicted in \cref{fig:datarecording}. The collected data from the proposed wearable sensing prototype had 10 channels per sensor, and a total of 20 channels: three channels (axes) of acceleration, three channels of gyroscope, three channels of magnetometer data, and one channel of body capacitance data. The data collected using the Apple Watch and iPhone mini had 9 channels per sensor, and a total of 27 channels: three channels of acceleration, three channels of gyroscope, and three channels of magnetometer data.
        We synchronized the recorded video data, left wrist sensor data, and right wrist sensor data by clapping five times at the start and end of each session. Based on the video data, we manually annotated the user's activities as 12 different activities, including a Null class. The proposed prototype sensor provided IMU sensor data and body capacitance data with a sampling rate of 25 Hz, while the Apple Watch provided only IMU sensor data with a sampling rate of 100 Hz. Overall, the experiment allowed us to evaluate the performance of the proposed sensor hardware and collect a dataset that can be used to develop and test algorithms for human intention recognition.

    \subsection{Dataset}

        \begin{table}[!t]
		\caption{Comparison of the data distribution of activities in the proposed sensor and Apple Watch datasets}
		\centering
		\label{tab:datadistribution}
		  \begin{tabular}{|p{0.1\columnwidth}|p{0.25\linewidth}|c|c|}
                \hline
                \multicolumn{2}{|c|}{} & Proposed    & Apple \\ \hline
                \multicolumn{2}{|c|}{\# of Subjects}  & 12    & 12    \\ \hline
                \multicolumn{2}{|c|}{\# of Session}  & 5    & 5    \\ \hline
                \multicolumn{2}{|c|}{\# of Channels per device}  & 10    & 9    \\ \hline
                \multicolumn{2}{|c|}{\# of Devices}  & 2    & 3    \\ \hline
                \multicolumn{2}{|c|}{Sampling frequency}  & 25 Hz    & 100 Hz    \\ \hline
                \multirow{12}{*}{Activity}  & Null  & 14m 9s (9.9\%)    & 29m 43s (17.5\%) \\ 
                &   Pressing button    & 2m 24s (1.7\%)        & 1m 54s (1.1\%) \\ 
                &   Sliding doorlock  & 1m 58s (1.4\%)        & 1m 21s (0.8\%) \\
                &   Opening door       & 8m 53s (6.2\%)        & 10m 53s (6.4\%)   \\
                &   Closing door      & 12m 3s (8.5\%)        & 14m 52s (8.7\%)   \\
                &   Checking machines  & 47m 6s (33.1\%)       & 48m 53s (28.7\%)  \\
                &   Walking         & 19m 13s (13.5\%)      & 32m 23s (19.0\%)  \\
                &   Taking key      & 4m 20s (3.0\%)        & 2m 21s (1.4\%)    \\
                &   Rotating key    & 10m 14s (7.2\%)       & 6m 47s (4.0\%)    \\
                &   Placing key back   & 6m 26s (4.5\%)     & 6m 16s (3.7\%)    \\
                &   Checking doorlock   & 2m 54s (2.0\%)    & 0m 29s (0.3\%)    \\
                &   Touching screen     & 12m 44s (8.9\%)   & 14m 11s (8.3\%)   \\
                \cline{2-4}
                &   Total           & 142m 24s              & 170m 3s \\
                \hline
            \end{tabular}
        \end{table}
        The collected sensor data by the proposed sensors and the Apple Watches were collected at different times. The total duration of the sensor data collected by the proposed sensors is 142 minutes and 24 seconds, while the total duration of the data collected by the Apple Watches is 170 minutes and 3 seconds. The goal of this work is to prevent collisions between workers and mobile robots, so we decided to collect more sensor data by the Apple Watches for the walking class after we collected the sensor data by the proposed sensor, resulting in a higher the proportion of the walking class data in the Apple Watches dataset compared to the proposed sensor data.
        
        To evaluate the proposed sensor hardware and neural network models for worker activity recognition, we annotated the user's activities using three different annotation schemes. The first scheme classified the sensor data into 12 activities, including Null, opening/closing doors, checking machines, walking, pressing buttons, and placing back keys. Since the human posture and locomotion are highly relevant in this scenario, we also used a second scheme that classified the data into four activities: Null, walking, working on upper parts of machines, and working on lower parts of machines. Lastly, since our main focus is on preventing collisions between workers and mobile robots, we also investigated a third annotation scheme that classified sensor data into two classes: walking and non-walking activities.
        
        % \begin{figure}[!t]
        %     \centering
        %     \subfloat[]{\includegraphics[width=0.45\columnwidth]{Figures/datadistribution_proposed.png}}
        %     \subfloat[]{\includegraphics[width=0.45\columnwidth]{Figures/datadistribution_proposed.png}}
        %     \caption{The data distribution of activities in (a) the proposed sensor dataset, (b) the Apple Watch dataset. }
        %     \label{fig:datadistribution}
        % \end{figure}

        \cref{tab:datadistribution} shows the distribution of activities in the proposed sensor dataset and the Apple Watch dataset. To facilitate processing with neural network models, we applied data augmentation using a sliding window of length 100 (1 second) and a step size of 4 (0.04 seconds) to the Apple Watch data, and a sliding window of length 25 (1 second) and a step size of 1 (0.04 seconds) to the proposed sensor data.

    \subsection{Neural Network Models for Worker Activity Recognition} % Sungho
        
        \begin{figure*}[!t]
            \centering
            \includegraphics[width=0.8\linewidth]{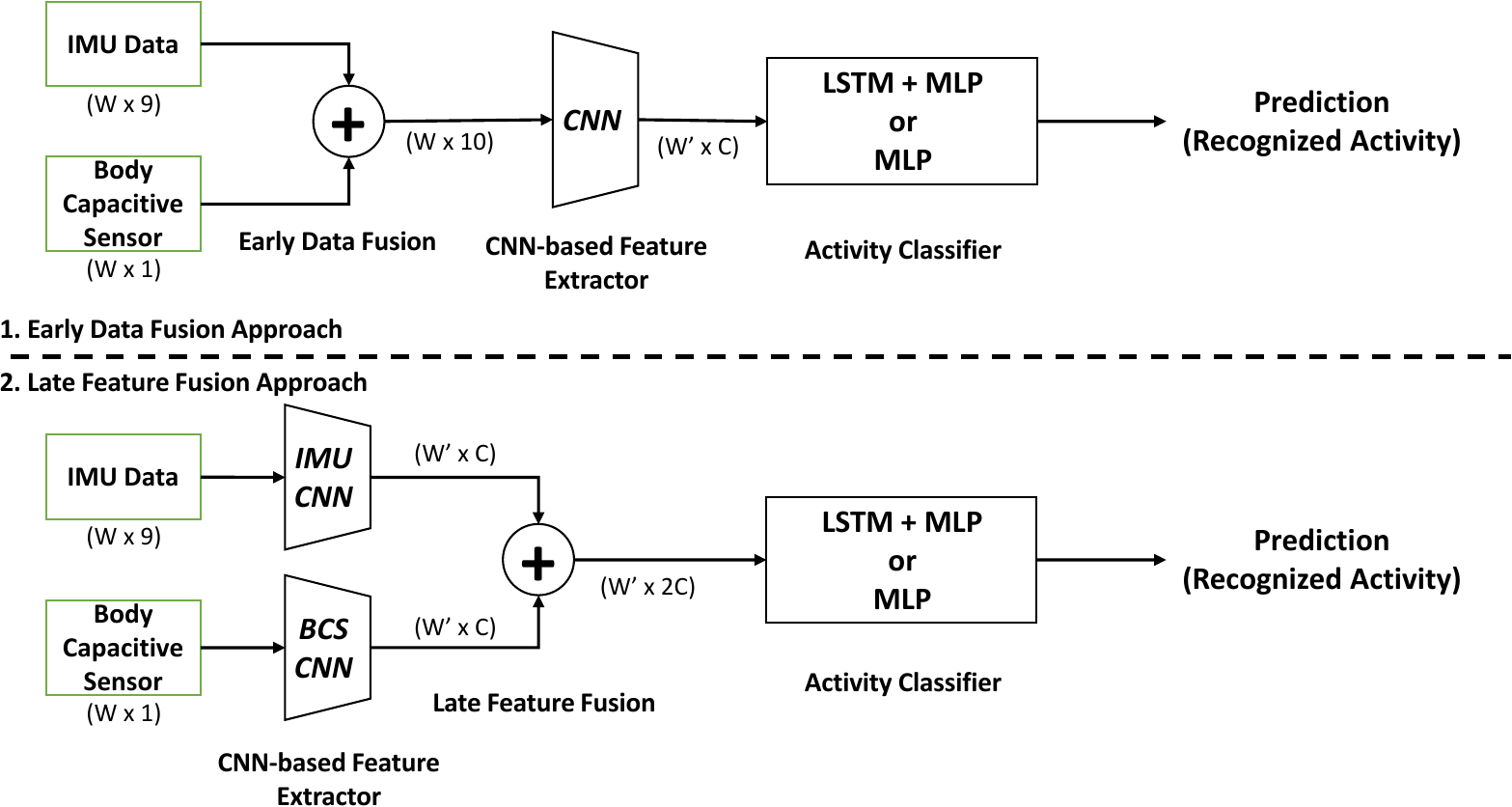}
            \caption{Overview of the proposed multimodal sensor fusion networks: Early data fusion and late feature fusion approaches. The activity classifiers of DeepConvLSTM and MC-CNN are LSTM with a Multilayer Perceptron (MLP) and an MLP, respectively. In the late feature fusion approach, two different CNNs, IMU-CNN and BCS-CNN, extract features from different sensor types.}
            \label{fig:multimodalfusion}
        \end{figure*}
        
        To evaluate the effectiveness of Apple Watch and the designed sensor, we used neural networks based on convolutional neural networks (CNN) \cite{sornam2017survey} and their combination with long short-term memory (LSTM) \cite{ordonez2016deep}. To train and validate the neural network models, we adopted a leave-one-session-out scheme. %For each recording, we perform the LOSO scheme meaning that f
        For each fold one session is used for the test, another for validation, and the remaining ones (3 sessions) for training. 
        
        The proposed sensor data contains two different types of data: the IMU data and the capacitive sensor data. To handle these multimodal sensor data, we propose and compare early and late sensor data fusion approaches for the multi-channel time-series convolutional neural networks (MC-CNN) and the deep convolutional LSTM (DeepConvLSTM). Early and late sensor data fusion approaches can be categorized depending on the position of the fusion within the processing chain \cite{gadzicki2020early}. Early data fusion gives all sensor data sources together as input to the neural network. Thus, the network extracts the unified features. If the multimodal sensor data are properly aligned, correlations between different sensor data may be exploited, thereby providing an opportunity to increase the performance of the system. Late feature fusion, on the other hand, merges features after separate full processing in independent feature extractor layers. The extracted features for each modality are concatenated and then go through further layers to predict the worker's activity class. 
        
        In this paper, we propose two different ensemble approaches for worker activity recognition: an early data fusion approach by merging the IMU and capacitive sensor data from the proposed wearable sensors to exploit correlations between different sensor data and improve the activity recognition performance, and a late feature fusion approach by concatenating an IMU sensor feature from the IMU feature extractor and a capacitive sensor feature from the capacitive sensor feature extractor to retain the characteristic of each sensor type and improve the activity recognition performance. 
        
        The overview of the early data fusion and late feature fusion approaches is depicted in \cref{fig:multimodalfusion}. The early data fusion approach merges the IMU and capacitive sensor data, while the late feature fusion approach concatenates the IMU sensor feature from the IMU-CNN feature extractor and the capacitive sensor feature from the body capacitive sensor (BCS)-CNN feature extractor.
        
    \subsection{Evaluation Metric and Implementation Details} 
    To evaluate the performance of the proposed hardware and neural network models, we adopted common evaluation metrics used in various human activity recognition studies, such as accuracy and macro F1 score \cite{ordonez2016deep, suh2022adversarial, suh2023tasked}. In addition, we specially reported the accuracy for the walking class, which is crucial for preventing the potential collision between worker and mobile robot.
	
    All experiments were implemented using Python scripts in the PyTorch framework on a Linux system with NVIDIA RTX A6000 GPUs. For the MC-CNN, we designed a feature extractor consisting of three convolutional layers with batch normalization, dropout, and max-pooling, and an activity classifier consisting of two fully connected layers. For the DeepConvLSTM, we used a feature extractor consisting of two convolutional layers with batch normalization, dropout, and max-pooling, and an activity classifier consisting of an LSTM layer and a fully connected layer. For all approaches, we trained the neural network models using Adam optimizer with a learning rate of $1 \times 10^{-4}$, a batch size of 128, and 200 training epochs. We employed early stopping with the patience of 20 epochs to prevent overfitting.
    
    \section{Experimental Results}
    \label{sec:ExpResults}
        \subsection{Results on the data acquired by the proposed sensors}
        % Comparison results with various deep learning models by changing the sensor types to emphasize the effect of the capacitive sensor (Ablation study?)
        
        \begin{table*}[!t]
		\caption{Comparison results of the proposed method with DeepConvLSTM and MC-CNN using proposed sensors with and without capacitive sensors in terms of the testing accuracy and Macro F1 score, depending on the number of annotated classes. The numbers are expressed in percent and represented as $mean \pm std$.}
		\centering
		\label{tab:ablationstudy}
		\begin{tabular}{|c|c|c|c|c|c|c|c|}
		    \hline
		        \multirow{3}{*}{\# of Classes}& \multirow{3}{*}{Metrics} & \multicolumn{4}{|c|}{With Capacitive Sensors}   & \multicolumn{2}{|c|}{Without Capacitive Sensors} \\
		        \cline{3-8}
		        & &  \multicolumn{2}{|c|}{DeepConvLSTM} & \multicolumn{2}{|c|}{MC-CNN}    & \multirow{2}{*}{DeepConvLSTM}  & \multirow{2}{*}{MC-CNN}\\
		        \cline{3-6}
		        & & Data Fusion & Feature Fusion & Data Fusion & Feature Fusion & & \\ 
		  %  \cline{2-8}
		    \hline
		    \multirow{3}{*}{12 Classes} & Accuracy    & 56.62 $\pm$ 2.69  & 58.20 $\pm$ 2.84  & 60.32 $\pm$ 3.29  & \textbf{62.70 $\pm$ 2.66} & 57.54 $\pm$ 2.60 & 58.82 $\pm$ 2.88 \\
		    & Macro F1    & 36.60 $\pm$ 1.64  & 41.80 $\pm$ 1.41  & 40.52 $\pm$ 2.22  & \textbf{46.91 $\pm$ 1.65} & 37.91 $\pm$ 1.32 & 41.63 $\pm$ 2.13 \\
		    & Walking Accuracy & 67.90 $\pm$ 7.17 & 69.52 $\pm$ 4.62 & 71.42 $\pm$ 2.62   &  \textbf{71.50 $\pm$ 2.35} & 68.82 $\pm$ 2.99 & 70.15 $\pm$ 8.08\\
		    \hline
		  %  \hline
		    \multirow{3}{*}{11 Classes} & Accuracy  & 63.87 $\pm$ 3.13  & 64.47 $\pm$ 2.65  & 63.76 $\pm$ 3.69  & \textbf{69.76 $\pm$ 3.09} & 63.45 $\pm$ 2.49 & 63.76 $\pm$ 3.90 \\
		    & Macro F1    & 39.52 $\pm$ 2.10    & 45.28 $\pm$ 1.36  & 44.23 $\pm$ 2.53  & \textbf{52.40 $\pm$ 2.32} & 40.24 $\pm$ 1.36 & 45.14 $\pm$ 3.38 \\
		    & Walking Accuracy & 72.91 $\pm$ 3.27   & \textbf{76.23 $\pm$ 3.22} & 69.76 $\pm$ 11.67 & 75.88 $\pm$ 2.55 & 69.84 $\pm$ 2.52 & 69.49 $\pm$ 7.14\\
		    \hline
		    \multirow{3}{*}{4 Classes}  & Accuracy  & 75.14 $\pm$ 1.46  & 76.19 $\pm$ 1.82  & \textbf{79.82 $\pm$ 0.56}  & 79.18 $\pm$ 0.89 & 78.54 $\pm$ 1.98 & 78.99 $\pm$ 1.32 \\
		    &   Macro F1    & 58.81 $\pm$ 1.93  & 60.01 $\pm$ 1.82  & 59.74 $\pm$ 1.93  & \textbf{63.97 $\pm$ 1.69} & 58.43 $\pm$ 2.52 & 58.90 $\pm$ 3.06 \\
		    &   Walking Accuracy & 61.22 $\pm$ 6.75 & 63.96 $\pm$ 6.51 & 58.91 $\pm$ 7.63   & \textbf{69.32 $\pm$ 4.26} & 68.70 $\pm$ 12.26 & 62.23 $\pm$ 14.88\\
		    \hline
		    \multirow{3}{*}{3 Classes}  & Accuracy  & 85.66 $\pm$ 1.68  & 86.67 $\pm$ 1.69  & 88.80 $\pm$ 1.10  & \textbf{88.94 $\pm$ 1.28}  & 85.19 $\pm$ 2.14  & 88.32 $\pm$ 1.20\\
		    &   Macro F1    & 77.11 $\pm$ 1.99  & 78.71 $\pm$ 2.27  & 82.06 $\pm$ 1.31  & \textbf{82.60 $\pm$ 1.55} & 76.76 $\pm$ 2.48  & 81.34 $\pm$ 1.63 \\
		    &   Walking Accuracy    & 68.67 $\pm$ 5.73 & 68.38 $\pm$ 4.88   & 70.30 $\pm$ 2.99  & \textbf{73.92 $\pm$ 3.86} & 67.84 $\pm$ 3.37  & 69.57 $\pm$ 3.77\\
		    \hline
		    \multirow{3}{*}{2 Classes}  & Accuracy    & 91.00 $\pm$ 0.60 & 91.57 $\pm$ 0.63  & 91.70 $\pm$ 0.56 & \textbf{91.85 $\pm$ 0.26} & 90.14 $\pm$ 0.95 & 91.30 $\pm$ 1.58 \\
		    &   Macro F1    & 78.94 $\pm$ 1.52 & 79.25 $\pm$ 1.70  & 79.63 $\pm$ 2.16   & \textbf{81.39 $\pm$ 0.97} & 77.43 $\pm$ 1.23 & 79.20 $\pm$ 2.06 \\
		    &   Walking Accuracy & 62.32 $\pm$ 7.10 & 58.62 $\pm$ 6.72 & 60.62 $\pm$ 11.08 & \textbf{68.43 $\pm$ 5.35} & 61.13 $\pm$ 7.30 & 60.51 $\pm$ 6.25\\
		    \hline
        \end{tabular}
    	\end{table*}
    	
    	In this subsection, we compare the performance of the proposed wearable sensors, both with and without capacitive sensors, using DeepConvLSTM and MC-CNN to demonstrate the effectiveness of the body capacitive sensor for worker activity recognition in the manufacturing line. In addition, we evaluate the sensor fusion approaches with MC-CNN and DeepConvLSTM using the proposed wearable sensor hardware that combines the IMU and body capacitive sensors. As we mentioned above, we annotated the user's activities using three different annotation schemes. To remove the impact of the Null class that can be confusing with other classes, we added two more annotation schemes removing the Null class data, such as 11 activities and 3 activities. \cref{tab:ablationstudy} shows the comparison results of the proposed method using the proposed sensors both with and without capacitive sensors, including five different annotation schemes. Data fusion means the early data fusion approach for DeepConvLSTM and MC-CNN, and Feature fusion means the late feature fusion approach for both neural network models. The comparison results demonstrate that the proposed wearable sensors with body capacitive sensors using the late feature fusion approach provided much better performance in terms of accuracy, macro F1 score, and walking class accuracy, than the sensors without body capacitive sensors and the sensors with body capacitive sensors using the early feature fusion approach Meanwhile, the proposed sensors with body capacitive sensors using the early data fusion approach showed similar performance compared the proposed sensors without body capacitive sensors, These results imply that the proposed late feature fusion can retain the characteristic of each sensor data and improve the activity recognition performance. In addition, the results on the schemes removing the Null class provided better performances than the schemes including the Null class, which further confirmed the impact of the Null class on the proposed wearable sensor data and neural networks models for worker activity recognition in the manufacturing line. Overall, the comparison results support the effectiveness of the proposed wearable sensors, particularly when the body capacitive sensor is incorporated with the late feature fusion approach.
    	
        \subsection{Comparison Results with Apple Watch Data}
        
        \begin{table*}[!t]
		\centering
		\caption{Comparison results of the proposed method with the Apple Watch sensor data with DeepConvLSTM and MC-CNN in terms of the testing accuracy and Macro F1 score by the number of annotated classes. The numbers are expressed in percent and represented as $mean \pm std$.}
		\label{tab:AppleResults}
		\begin{tabular}{|c|c|c|c|c|c|c|c|}
		    \hline
		        \multirow{2}{*}{\# of Classes}& \multirow{2}{*}{Metrics}&  \multicolumn{2}{|c|}{2$\times$ Apple Watch and 1$\times$iPhone}   & \multicolumn{2}{|c|}{2$\times$ Apple Watch} & \multicolumn{2}{|c|}{Ours}\\
		        \cline{3-8}
		        & &  DeepConvLSTM & MC-CNN    & DeepConvLSTM  & MC-CNN  & DeepConvLSTM  & MC-CNN \\
		    \hline
		    \multirow{3}{*}{12 Classes} & Accuracy    & 71.07 $\pm$ 2.14  & \textbf{72.80 $\pm$ 2.80} & 57.46 $\pm$ 2.71 & 61.33 $\pm$ 2.64 & 58.20 $\pm$ 2.84    & 62.70 $\pm$ 2.66\\
		    &   Macro F1    & 47.03 $\pm$ 1.73  & \textbf{48.51 $\pm$ 3.22} & 38.23 $\pm$ 1.63 & 40.71 $\pm$ 2.08 & 41.80 $\pm$ 1.41    & 46.91 $\pm$ 1.65\\
		    & Walking Accuracy & \textbf{75.21 $\pm$ 4.98} & 69.15 $\pm$ 5.82 & 62.23 $\pm$ 5.50 & 62.12 $\pm$ 8.09 & 69.52 $\pm$ 4.62    & 71.50 $\pm$ 2.35\\
		    \hline
		    \multirow{3}{*}{11 Classes} & Accuracy    & 63.63 $\pm$ 2.22 & 67.94 $\pm$ 2.55 & 67.35 $\pm$ 2.15 & 69.49 $\pm$ 2.74 & 64.47 $\pm$ 2.65 & \textbf{69.76 $\pm$ 3.09}\\
		    & Macro F1    & 39.73 $\pm$ 1.69 & 43.02 $\pm$ 2.49 & 44.10 $\pm$ 1.56 & 44.72 $\pm$ 1.80  & 45.25 $\pm$ 1.36  & \textbf{52.40 $\pm$ 2.32} \\
		    & Walking Accuracy & 74.63 $\pm$ 4.86 & 69.96 $\pm$ 4.39 & 71.86 $\pm$ 5.65 & 72.12 $\pm$ 8.34 & \textbf{76.23 $\pm$ 3.22}  & 75.88 $\pm$ 2.55\\
		    \hline
		    \multirow{3}{*}{4 Classes}  & Accuracy    & 84.31 $\pm$ 1.60 & \textbf{84.65 $\pm$ 2.46} & 67.58 $\pm$ 2.41 & 70.70 $\pm$ 3.73 & 76.16 $\pm$ 1.82   & 79.18 $\pm$ 0.89\\
		    & Macro F1    & \textbf{84.64 $\pm$ 1.37} & 84.29 $\pm$ 2.43   & 62.70 $\pm$ 2.11 & 64.01 $\pm$ 3.38    & 60.01 $\pm$ 1.82  & 63.97 $\pm$ 1.69\\
		    & Walking Accuracy & \textbf{71.99 $\pm$ 4.90}     & 62.37 $\pm$ 7.40  & 68.11 $\pm$ 4.99  & 54.20 $\pm$ 7.79 & 63.96 $\pm$ 6.51    & 69.32 $\pm$ 4.26\\
		    \hline
		    \multirow{3}{*}{3 Classes}  & Accuracy  & 78.58 $\pm$ 1.39  & 81.71 $\pm$ 1.32  & 79.35 $\pm$ 1.98  & 82.40 $\pm$ 2.05  & 86.67 $\pm$ 1.69  & \textbf{88.94 $\pm$ 1.28} \\
		    & Macro F1  & 76.81 $\pm$ 1.25  & 80.19 $\pm$ 1.15  & 77.62 $\pm$ 1.97  & 80.88 $\pm$ 2.04  & 78.71 $\pm$ 2.27  & \textbf{82.60 $\pm$ 1.55}\\
		    & Walking Accuracy  & 70.69 $\pm$ 4.72  & 72.19 $\pm$ 4.83  & 70.34 $\pm$ 5.47  & 73.18 $\pm$ 5.15  & 68.38 $\pm$ 4.88  & \textbf{73.92 $\pm$ 3.86}\\
		    \hline
		    \multirow{3}{*}{2 Classes}  & Accuracy    & 90.25 $\pm$ 1.05  & \textbf{92.54 $\pm$ 1.25}  & 89.46 $\pm$ 2.07  & 92.03 $\pm$ 1.90 & 91.57 $\pm$ 0.63 & 91.85 $\pm$ 0.26 \\
		    & Macro F1    & 85.35 $\pm$ 1.29  & \textbf{85.84 $\pm$ 2.83}  & 81.80 $\pm$ 2.85  & 85.01 $\pm$ 3.71 & 79.25 $\pm$ 1.70    & 81.39 $\pm$ 0.97\\
		    & Walking Accuracy & \textbf{72.88 $\pm$ 5.62} & 53.76 $\pm$ 8.83  & 67.62 $\pm$ 5.76  & 59.04 $\pm$ 7.49   & 58.62 $\pm$ 6.72  & 68.43 $\pm$ 5.35\\
		    \hline
        \end{tabular}
    	\end{table*}
    	
    	% We can show the performance difference between the designed sensor and Apple Watch to reveal the advantage of the capacitive sensor
    	In this subsection, we compare the proposed sensor framework with the Apple Watch to highlight the advantages of the body capacitive sensor. As mentioned earlier, the data collected by the proposed sensors and the Apple Watch were recorded at different times, and the activity distributions in the datasets were different. Using MC-CNN and DeepConvLSTM, we evaluated the Apple Watch data with five annotation schemes identical to the previous subsection. To ensure fair comparisons, we also processed the Apple Watch data without the iPhone mini data acquired from the left arm, since the proposed sensors were worn on only both wrists. The comparison results of the proposed method with the Apple Watch sensor data, including accuracy, macro F1 score, and walking class accuracy using five annotation schemes, are presented in Table \ref{tab:AppleResults}. Despite the lower sampling rate of 25 Hz for the proposed sensor data compared to the 100 Hz Apple Watch sampling rate, our proposed method outperformed the Apple Watch on both wrists without the iPhone mini on the left arm in terms of accuracy, macro F1 score, and walking class accuracy. 
    	
        Interestingly, we observed that the results of the Apple Watch sensor data with the iPhone mini only improved the performance of the activity recognition in annotation schemes including Null class, while the performance in schemes without Null class, such as 11 classes, 3 classes, and 2 classes, was similar to that of the Apple Watch sensor data without the iPhone mini. This suggests that the iPhone mini on the left arm may have an impact on distinguishing between the Null class and other classes, rather than between normal activities.
        
        % \subsection{Discussion}
        
        % To further improve the performance of the activity recognition, we plan to collect more data from more volunteers and apply generative adversarial networks (GAN) \cite{wang2018sensorygans} and domain generalization methods \cite{soleimani2021cross, suh2021adversarial}. If the performance of the activity recognition by using sensor data only is not quite good, we will combine the sensor data with video data and feed it into the neural networks. 

        % The feasibility of developing machine learning models for human activity recognition was validated through the development of an adversarial encoder-decoder structure with maximum mean discrepancy to realign the data distribution over multiple subjects, and tested on four open datasets. \cite{suh2022adversarial} report that the results obtained outperformed state-of-the-art methods and improved generalization capabilities. The same authors also proposed TASKED (Transformer-based Adversarial learning framework for human activity recognition using wearable sensors via Self-KnowledgE Distillation), a deep learing architecture capable of learning cross-domain feature representations using adversarial learning and maximum mean discrepancy to align data distributions from multiple domains \cite{suh2023tasked}. Future work will address the development of models for human intention recognition and how such predictions can be leveraged for safe zone detection when routing autonomous mobile robots in the manufacturing context.
        
    \section{Conclusion}
    \label{sec:conclusion}
    
        % Statement of the proposed hardware design and the use case
        % Limitations and Future work
        
        In this paper, we proposed a novel wearable sensing prototype that combines IMU and body capacitance sensing module to recognize worker activities in the manufacturing line. To handle these multimodal sensor data, we propose and compare early and late sensor data fusion approaches for MC-CNN and DeepConvLSTM neural networks. We demonstrate the effectiveness of the proposed wearable sensors and neural network models through a series of experiments on the collected datasets. Our experimental results show that the proposed prototype can achieve high recognition accuracy in recognizing worker activities in the manufacturing line. We also show that the late fusion approach outperformed the early fusion approach for the proposed wearable sensor data with the IMU and body capacitive sensors, and the proposed wearable sensors provided better performance than the Apple Watch though the lower sampling frequency of 25 Hz for the proposed sensor data compared to the 100 Hz sampling frequency of the Apple Watch data. The proposed sensing prototype with a body capacitive sensor and feature fusion method improves by 6.35\% and yields a 9.38\% higher macro F1 score than the proposed sensing prototype without a body capacitive sensor and Apple Watch data, respectively

        In the future, we plan to explore more advanced feature extraction and data fusion approaches to further improve the recognition accuracy of worker activities. We also plan to investigate the generalizability of the proposed prototype and neural network models to different manufacturing settings and worker populations. Additionally, we aim to collect more data from more volunteers and apply generative adversarial networks (GAN) \cite{wang2018sensorygans} and domain generalization methods \cite{suh2022adversarial} to further enhance the performance of the activity recognition. Additionally, we will explore the combination of sensor data with video data to improve activity recognition. 
        Moreover, we will also focus on the development of models for human intention recognition and how such predictions can be leveraged for safe zone detection when routing autonomous mobile robots in the manufacturing context.

% \section{Tables}
% Note that, for IEEE-style tables, the
%  $\backslash${\tt{caption}} command should come BEFORE the table. Table captions use title case. Articles (a, an, the), coordinating conjunctions (and, but, for, or, nor), and most short prepositions are lowercase unless they are the first or last word. Table text will default to $\backslash${\tt{footnotesize}} as
%  the IEEE normally uses this smaller font for tables.
%  The $\backslash${\tt{label}} must come after $\backslash${\tt{caption}} as always.
 
% \begin{table}[!t]
% \caption{An Example of a Table\label{tab:table1}}
% \centering
% \begin{tabular}{|c||c|}
% \hline
% One & Two\\
% \hline
% Three & Four\\
% \hline
% \end{tabular}
% \end{table}

\section*{Compliance with Ethical Standards}
% This research study was performed in line with the principles of the Declaration of Helsinki, and utilized human sensor data that was collected from experiment participants. The participants gave informed consent in accordance with the policies of the Ethics Team of the German Research Center for Artificial Intelligence (DFKI), which approved the experimental protocol.
This research study was performed in line with regional ethical principles, including the Declaration of Helsinki and local policies as required by the German Research Center for Artificial Intelligence (DFKI).

\section*{Acknowledgments}
This work was supported by the European Union’s Horizon 2020 program projects STAR under grant agreements numbers H2020-956573.
The Carl-Zeiss Stiftung also funded it under the Sustainable Embedded AI project (P2021-02-009).

% {\appendix[Proof of the Zonklar Equations]
% Use $\backslash${\tt{appendix}} if you have a single appendix:
% Do not use $\backslash${\tt{section}} anymore after $\backslash${\tt{appendix}}, only $\backslash${\tt{section*}}.
% If you have multiple appendixes use $\backslash${\tt{appendices}} then use $\backslash${\tt{section}} to start each appendix.
% You must declare a $\backslash${\tt{section}} before using any $\backslash${\tt{subsection}} or using $\backslash${\tt{label}} ($\backslash${\tt{appendices}} by itself
%  starts a section numbered zero.)}

%{\appendices
%\section*{Proof of the First Zonklar Equation}
%Appendix one text goes here.
% You can choose not to have a title for an appendix if you want by leaving the argument blank
%\section*{Proof of the Second Zonklar Equation}
%Appendix two text goes here.}

\bibliographystyle{IEEEtran}
\bibliography{ref}

%\newpage

% \section{Biography Section}
% If you have an EPS/PDF photo (graphicx package needed), extra braces are
%  needed around the contents of the optional argument to biography to prevent
%  the LaTeX parser from getting confused when it sees the complicated
%  $\backslash${\tt{includegraphics}} command within an optional argument. (You can create
%  your own custom macro containing the $\backslash${\tt{includegraphics}} command to make things
%  simpler here.)
\vspace{-10mm}

    \begin{IEEEbiography}[{\includegraphics[width=1in, height=1.25in, clip,keepaspectratio]{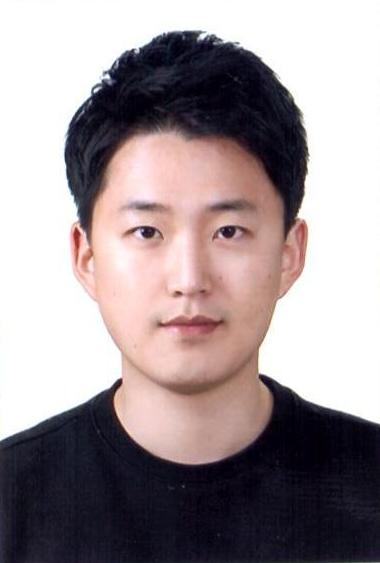}}]{Sungho Suh}
		is a Senior Researcher at the German Research Center for Artificial Intelligence (DFKI) in Germany since 2020. He received the Ph.D. degree in Computer Science at the Technische Universit{\"a}t Kaiserslautern, Germany in 2021, and the B.S. and M.S. degrees from the School of Electrical Engineering and Computer Science, Seoul National University, Seoul, South Korea, in 2009 and 2011, respectively. Before joining DFKI, he worked at KIST Europe in Germany for three years, and at Samsung Electro-Mechanics, Korea from 2011 to 2018. His research interests are machine learning algorithms, such as sensor data processing, computer vision, multimodal processing, and generative model, with a focus on industrial applications.
    \end{IEEEbiography}
\vspace{-10mm} 
    
    \begin{IEEEbiography}[{\includegraphics[width=1in, height=1.25in, clip,keepaspectratio]{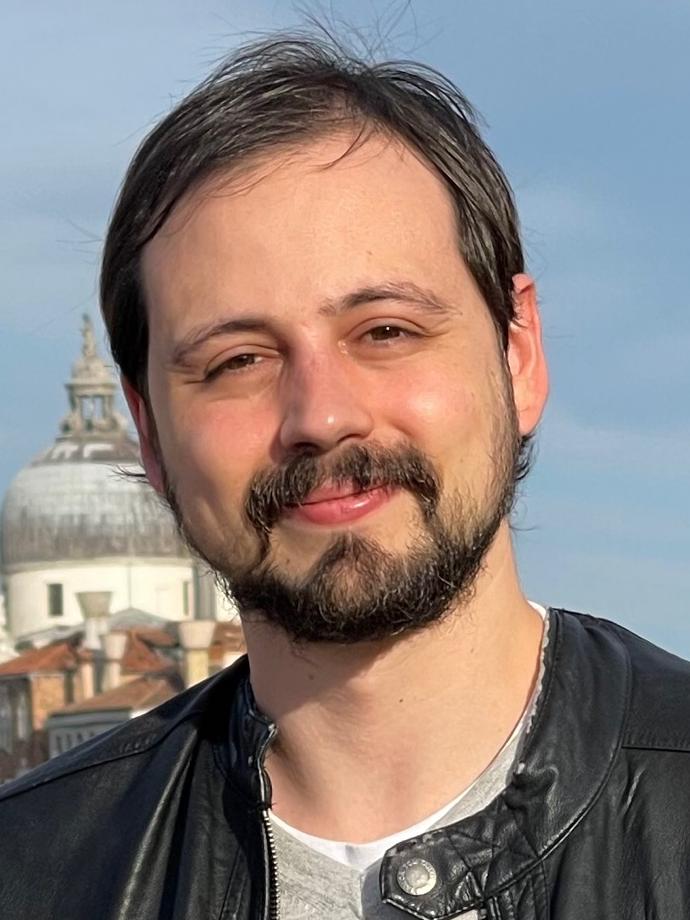}}]{Vitor Fortes Rey}
        		is a Researcher at the German Research Center for Artificial Intelligence (DFKI) in Germany since 2019. He is a Ph.D. candidate in Computer Science at the RPTU Kaiserslautern-Landau, Germany, and the B.S. and M.S. degrees from the Universidade Federal do Rio Grande do Sul, Porto Alegre, Brazil, in 2013 and 2016, respectively. His research interests are machine learning, multi-modal human activity recognition, and knowledge engineering.
    \end{IEEEbiography}
\vspace{-10mm} 
    
    \begin{IEEEbiography}[{\includegraphics[width=1in, height=1.25in, clip, keepaspectratio]{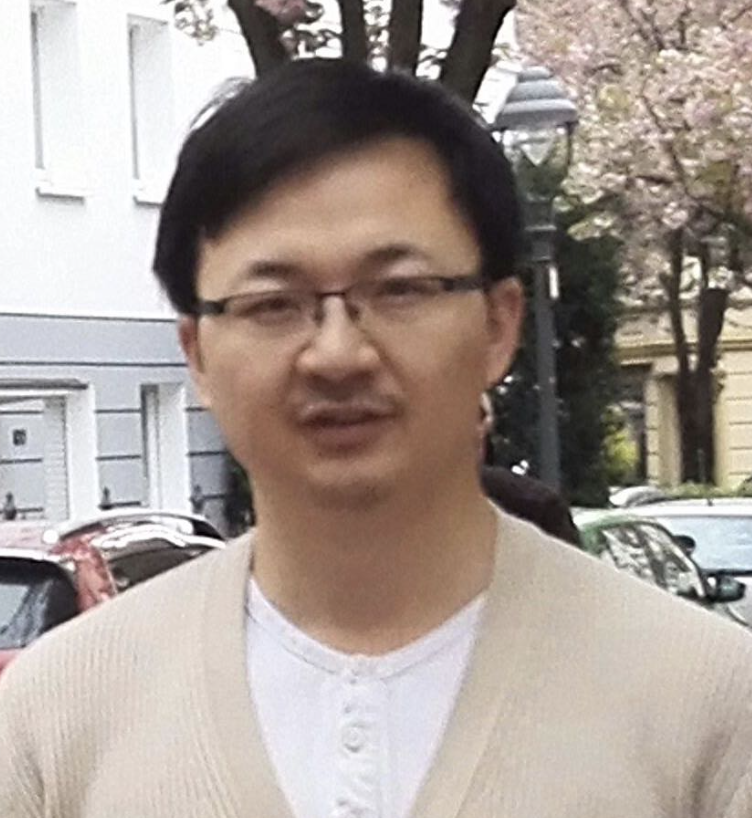}}]{Sizhen Bian}
		is a Senior Researcher at ETH Zürich in Switzerland since 2022. He received the Ph.D. degree in Computer Science at the Technische Universit{\"a}t Kaiserslautern, Germany in 2022, and the B.S. and M.S. degrees from the Northwestern Polytechnical University, China, and the Technische Universit{\"a}t Kaiserslautern, Germany, in 2013 and 2016, respectively. His research interests are spiking neural networks, pervasive computing, and efficient edge computing.
    \end{IEEEbiography}
    % \begin{IEEEbiography}[{\includegraphics[width=1in, height=1.25in, clip,keepaspectratio]{biography/sizhen.jpg}}]{Sizhen Bian}
    % \end{IEEEbiography}
    
    \begin{IEEEbiography}[{\includegraphics[width=1in, height=1.25in, clip,keepaspectratio]{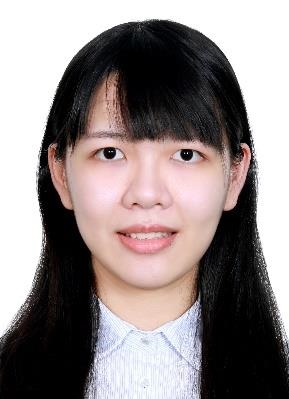}}]{Yu-Chi Huang}
        is a master student at the RPTU Kaiserslautern-Landau, Germany, and a research assistant at the German Research Center for Artificial Intelligence (DFKI). Her research interests are machine learning and computer vision techniques.
    \end{IEEEbiography}
    
    \begin{IEEEbiography}[{\includegraphics[width=1in, height=1.25in, clip,keepaspectratio]{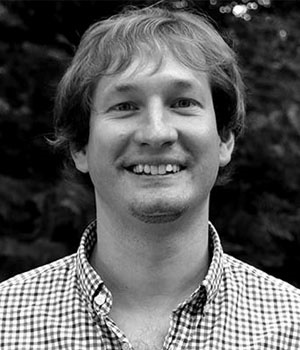}}]{Jože Martin Rožanec}
        is a Ph.D. candidate in Information and Communication Technologies at Jožef Stefan International Postgraduate School. He is a Researcher at the Artificial Intelligence Laboratory (Jožef Stefan Institute), and a machine learning engineer at Qlector d.o.o. (developing intelligent solutions for smart factories). He collaborates with the American Slovenian Education Foundation, where he leads multiple activities for Fellows and Alumni. Over more than ten years, he worked for several companies (e.g., Mercado Libre, Navent, Globant) in software engineering and machine learning-related roles. His research interests include machine learning methods for recommendations, fraud detection, demand forecasting, active learning, and explainable artificial intelligence (XAI).
    \end{IEEEbiography}
    
    \begin{IEEEbiography}[{\includegraphics[width=1in, height=1.25in, clip,keepaspectratio]{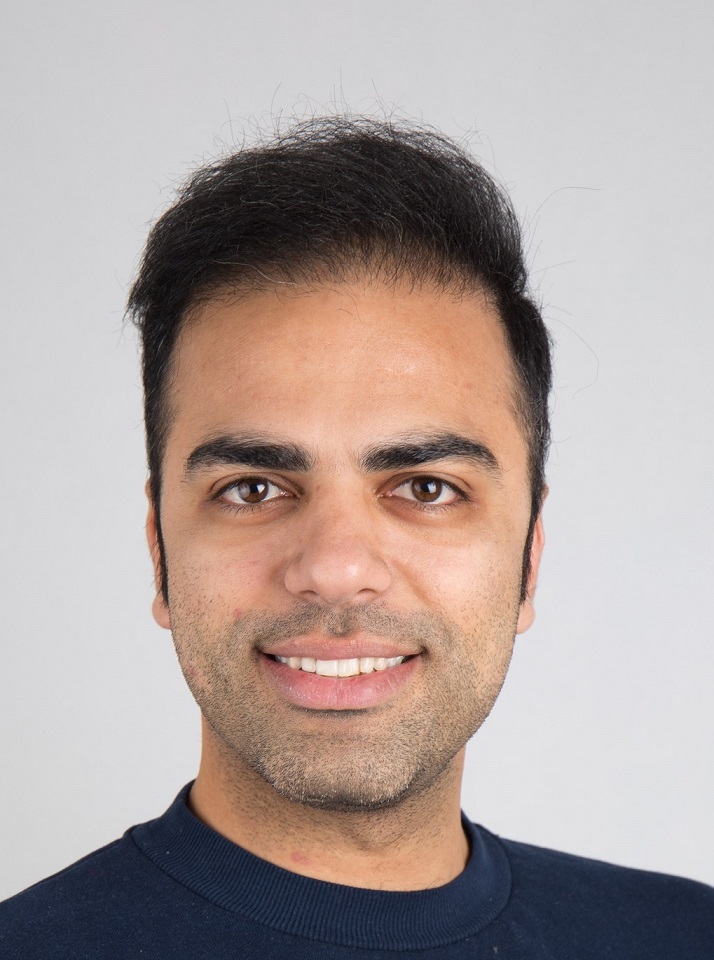}}]{Hooman Tavakoli Ghinani} is a project manager and Ph.D. student in SmartFactory-DFKI, Germany, obtained his bachelor's degree from the computer science department at Azad University in Najafabad, Iran. He then pursued a master's degree in intelligent systems at the computer science department of the Technical University of Kaiserslautern, Germany, which he completed in 2020. His master's thesis focused on 'Extensive Study of Probabilistic Regression Using GANs'. His research interests include continual learning in artificial intelligence systems, GANs, and vision-based human-machine interaction.
    \end{IEEEbiography}
    
    \begin{IEEEbiography}[{\includegraphics[width=1in, height=1.25in, clip,keepaspectratio]{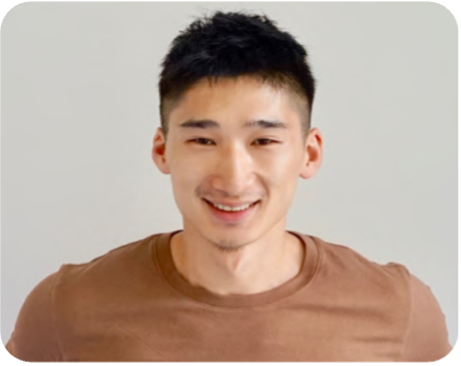}}]{Bo Zhou}
    is a Senior Researcher at the German Research Center for Artificial Intelligence (DFKI GmbH). His main research interests are sensor-based perception through hardware-software co-designed systems and sensor signal analysis. He received his Ph.D. in Computer Science at RPTU Kaiserslautern-Landau, and his M.S. degrees in Electronics from the University of Southampton, RPTU Kaiserslautern-Landau and NTNU.
    \end{IEEEbiography}
    
	\begin{IEEEbiography}[{\includegraphics[width=1in, height=1.25in, clip, keepaspectratio]{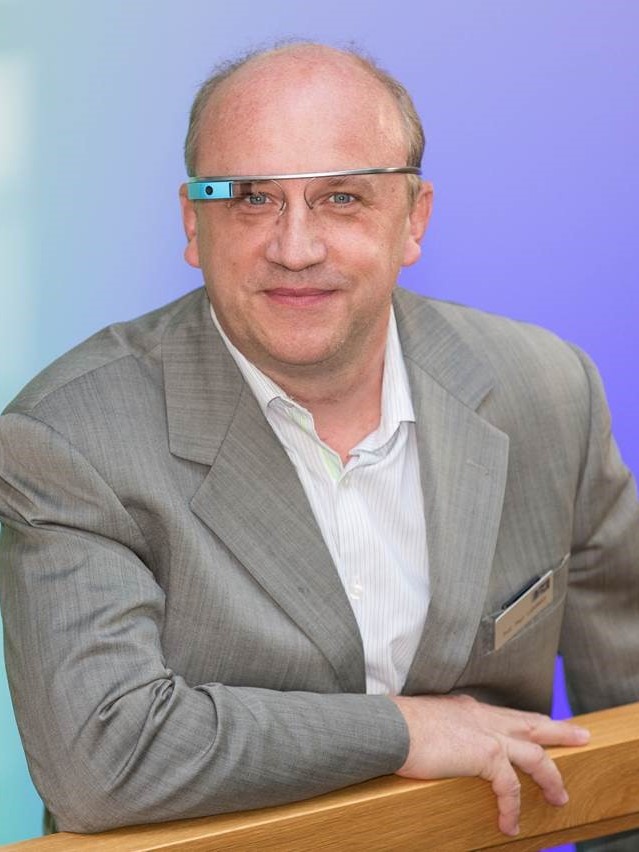}}]{Paul Lukowicz}
		is both Scientific Director at the German Research Center for Artificial Intelligence (DFKI GmbH) and Professor of Computer Science at RPTU Kaiserslautern-Landau in Germany since 2012 where he heads the Embedded Intelligence group. His research focuses on context-aware ubiquitous and wearable systems including sensing, pattern recognition, system architectures, models of large-scale self-organized systems, and applications in areas ranging from healthcare through industry 4.0 to smart cities.
	\end{IEEEbiography}

\vfill

\end{document}